\documentclass[runningheads]{llncs}

\usepackage{eccv}

\usepackage{eccvabbrv}
\usepackage{bm}
\usepackage{graphicx}
\usepackage{booktabs}
\usepackage{amsmath}
\usepackage{multirow}
\usepackage[accsupp]{axessibility}  
\renewcommand{\footnoterule}{%
  \kern 3pt 
  \hrule width 100pt height 0.5pt 
  \kern 2pt 
}

\usepackage[pagebackref,breaklinks,colorlinks,citecolor=eccvblue]{hyperref}
\usepackage{orcidlink}

\begin{document}

\title{Beyond Prompt Learning: Continual Adapter for Efficient Rehearsal-Free Continual Learning} 

\titlerunning{Continual Adapter}

\author{Xinyuan Gao\inst{1} \and
Songlin Dong\inst{2} \and
Yuhang He\inst{2} \and
Qiang Wang\inst{1} \and \\
Yihong Gong\inst{1,2}}

\authorrunning{X.~Gao et al.}

\institute{School of Software Engineering, Xi’an Jiaotong University \and
College of Artificial Intelligence, Xi’an Jiaotong University\\
\email{\{gxy010317,dsl97273141,hyh13794787,qwang\}@stu.xjtu.edu.cn, \\ ygong@mail.xjtu.edu.cn}}

\maketitle

\begin{abstract}
The problem of Rehearsal-Free Continual Learning~(RFCL) aims to continually learn new knowledge while preventing forgetting of the old knowledge, without storing any old samples and prototypes. The latest methods leverage large-scale pre-trained models as the backbone and use key-query matching to generate trainable prompts to learn new knowledge. However, the domain gap between the pre-training dataset and the downstream datasets can easily lead to inaccuracies in key-query matching prompt selection when directly generating queries using the pre-trained model, which hampers learning new knowledge. Thus, in this paper, we propose a \emph{beyond prompt learning} approach to the RFCL task, called Continual Adapter (C-ADA). It mainly comprises a parameter-extensible continual adapter layer (CAL) and a scaling and shifting~(S\&S) module in parallel with the pre-trained model. C-ADA flexibly extends specific weights in CAL to learn new knowledge for each task and freezes old weights to preserve prior knowledge, thereby avoiding matching errors and operational inefficiencies introduced by key-query matching. To reduce the gap, C-ADA employs an S\&S module to transfer the feature space from pre-trained datasets to downstream datasets. Moreover, we propose an orthogonal loss to mitigate the interaction between old and new knowledge. Our approach achieves significantly improved performance and training speed, outperforming the current state-of-the-art (SOTA) method. Additionally, we conduct experiments on domain-incremental learning, surpassing the SOTA, and demonstrating the generality of our approach in different settings.

\end{abstract}
\section{Introduction}
\label{sec:intro}

\begin{figure}[t]
\setlength{\abovecaptionskip}{-0.15cm} 
\setlength{\belowcaptionskip}{-0.5cm} 
\begin{center}
    \includegraphics[width=1.0\textwidth]{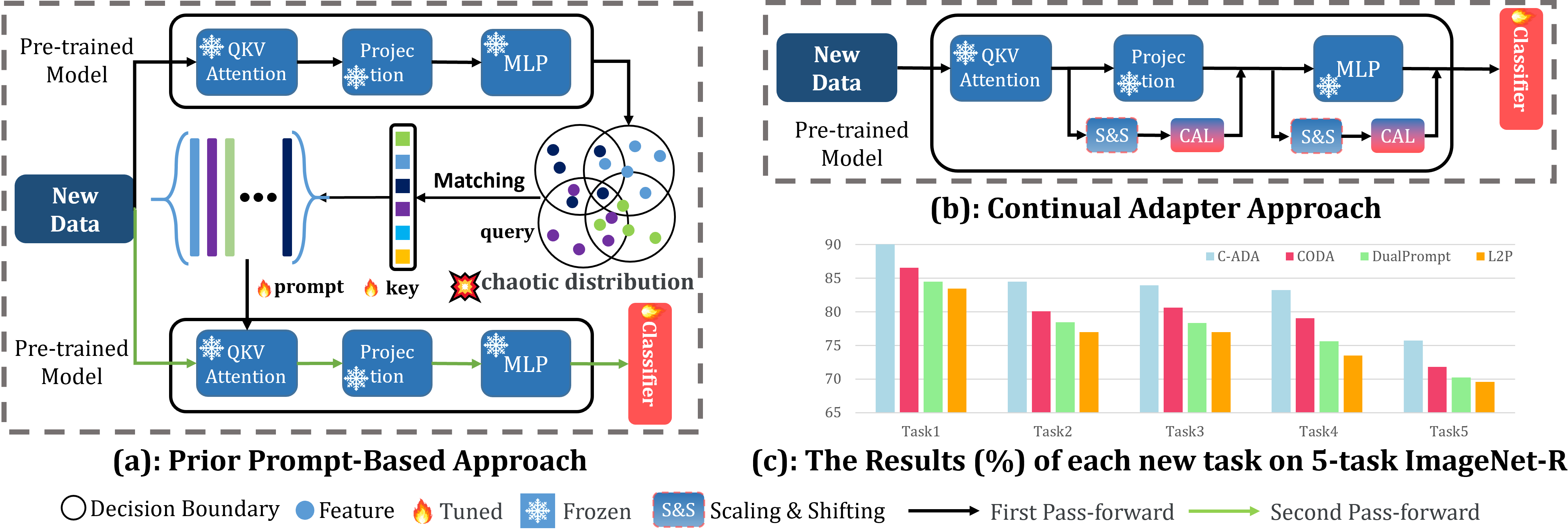}
\end{center}

\caption{\textbf{Prior prompt-based approaches} pass through the pre-trained model to generate the query and employ key-query matching to select prompts, which are inserted into the pre-trained model again~(each layer has a unique prompt). \textbf{Continual Adapter approach~(C-ADA)} strategy eliminates the need for key-query matching by introducing the CAL and S\&S, \textit{which brings significant improvement in learning the new knowledge}. Moreover, C-ADA only needs to pass the pre-trained model once, highlighting the training speed.}
\label{fig:head}

\end{figure}

In recent years, with the emerging applications of deploying DNN models to dynamic and continuous environments such as autonomous driving\cite{verwimp2023clad,shieh2020continual}. Continual Learning (CL)~\cite{ICARL,lucir,l2p,smith2023coda,der} has drawn a rapidly increasing research interest both in the literature and industry. 
Classical CL methods focus on addressing the problem of \emph{catastrophic forgetting}\cite{CF}, wherein the models suffer dramatic performance decreases on previous tasks once training on new data. The majority of research studies\cite{ICARL,tao2020topology,lucir,dong2021few,gao2023dkt,der,zhu2021prototype,zhu2022self} attempt to reduce forgetting by retaining a subset of previous image samples or prototypes and replaying them in new tasks, referred to as the \textbf{rehearsal-based methods}. However, on the concerns of data privacy and deployment efficiency, the applications of these methods in real-world scenarios are hampered. Therefore, we focus on a more challenging \textbf{rehearsal-free CL} (RFCL)\cite{LWF,EWC,l2p,dualprompt,smith2023coda}, which does not store any previous samples or prototypes.

Latest approaches\cite{l2p,smith2023coda,dualprompt, tang2023prompt} leverage prompt learning techniques to solve the RFCL problem. Due to freezing the large-scale pre-trained model, \textbf{learning new knowledge (\emph{i.e. plasticity}) becomes more challenging rather than reducing the forgetting (\emph{i.e. stability})\cite{smith2023coda}}. 
To acquire new knowledge, prompt-based methods utilize \emph{key-query matching} to obtain trainable prompts. As depicted in Figure \ref{fig:head}(a), these methods first generate a feature representation (\emph{i.e. query}) for each instance using the pre-trained model. Subsequently, they select or generate prompts for each instance based on the cosine similarity between the query and keys, followed by a second pass through the pre-trained model. Despite their promising progress, they are faced with two challenges. 1) Due to the substantial gap between the pre-trained dataset and the downstream datasets, the query embeddings directly generated by the pre-trained model are prone to confusion, leading to inaccurate key-query matching prompt selection. The inaccurate selection seriously hampers the learning ability of the model. 2) These methods require two forward passes, one for calculating the query and another for logit, leading to relatively low efficiency. Therefore, there are inherent shortcomings in prompt-based methods for RFCL tasks, motivating us to explore a novel approach distinct from prompt-learning techniques.

Taking into account the inherent drawbacks mentioned above, in this paper, we shift our focus to adapter tuning~\cite{houlsby2019parameter} and propose a simple yet effective adapter variant, named \textbf{Continual Adapter (C-ADA)} for rehearsal-free continual learning. Compared to prompt-learning techniques, C-ADA \emph{eliminates} the necessity for key-query matching, fundamentally avoiding issues arising from erroneous key-query matches that could lead to conflicts in training parameters. As depicted in Figure~\ref{fig:head}~(c), this results in a substantial enhancement in the accuracy of C-ADA for new tasks, and C-ADA achieves this with just a \emph{single forward pass}, greatly boosting training speed. Moreover, in contrast to \emph{lacking incremental capability}  adapter fine-tuning methods \cite{houlsby2019parameter,chen2022adaptformer}, we replace the adapter layer with a \textit{plug-and-play, parameter-expandable adapter layer}, named Continuous Adapter Layer (CAL). The CAL expands the specific learnable weights in a new dimension (middle dimension) to acquire new knowledge and freeze the previously learned weights to retain old knowledge, which empowers the model with the capacity to tackle complex continual scenarios.
This fundamentally encourages knowledge reuse, as the previous projection weights will contribute to future projection weights, facilitating the learning of new knowledge. We also introduce an orthogonal loss function to alleviate the conflict between new weights and previous weights, which significantly reduces the forgetting of old tasks. To further solve the problem of the gap between pre-trained datasets and downstream datasets, C-ADA includes scaling and shifting~(S\&S) module with negligible parameters to transfer the feature space from pre-trained datasets to downstream datasets before the incremental phase. Compared to the current SOTA methods, our model demonstrates significant improvements in both performance and training speed under equivalent parameter counts for RFCL. Furthermore, to validate its generality, we extend the method to the Domain-incremental Learning (DIL) setting and once again outperform the SOTA methods.
\emph{The key contributions of this paper are summarized as follows:}
\begin{itemize}
\item We propose a simple yet effective approach called C-ADA, which eliminates the necessity for key-query matching, for RFCL, characterized by \textit{better plasticity and more efficient training} compared to existing continual prompting approaches.

\item We design a novel CAL module, which comprises a parameter-extensible down-projection and up-projection. It facilitates the preservation of previously acquired knowledge and the learning of new knowledge. 

\item We develop an S\&S module to further reduce the divergence between the pre-training dataset and the downstream datasets, thereby enhancing the learning of new tasks. Additionally, we introduce an orthogonal loss to mitigate the interaction between old and new knowledge, which significantly contributes to the preservation of old knowledge.

\item We conduct extensive experiments on well-established rehearsal-free benchmarks and achieve a new SOTA performance. Meanwhile, we evaluate our approach on the DIL setting and again outperform the state of the art, highlighting the generality.
\end{itemize}
\section{Related work}
\subsection{Rehearsal-Based Continual Learning} \cite{tao2020topology,der,dytox,wang2022foster,gao2023dkt,zhu2021prototype,zhu2022self,wang2022s,dong2021few, wang2023non, wang2023semantic, song2024non} retain prior information~(old exemplars or prototypes) for subsequent tasks, aimed at mitigating catastrophic forgetting by replay them in the new tasks. \textit{Exemplars-based methods}~\cite{tao2020topology,dong2021few,der,dytox,wang2022foster,gao2023dkt} store a portion of old exemplars in a memory buffer. These exemplars are trained alongside the current samples to optimize the model. For instance, iCaRL~\cite{ICARL} and its variants~\cite{tao2020topology,lucir,der} prevent forgetting by employing the herding technique for exemplar selection and designing distinct distillation losses. However, exemplar-based methods not only require extra memory but also experience a significant performance decline as the buffer size decreases.
\textit{Exemplars-free methods}~\cite{zhu2021prototype,zhu2022self} typically store a prototype per class to refine the classifier. For example, \cite{zhu2021prototype} memorizes and augments prototypes to maintain the decision boundary.
Some \emph{latest} methods SLCA\cite{zhang2023slca}, RanPAC\cite{mcdonnell2024ranpac}, HiDe-Prompt\cite{wang2024hierarchical} combine the pre-trained model and \emph{prototypes} to achieve better performance. Since they are \emph{not} rehearsal-free continual learning, we have \emph{excluded} comparisons with them.

Nevertheless, preserving previous information may introduce privacy and data leakage risks, often impractical in real-world applications. This motivates us to concentrate on the \textbf{rehearsal-free} setting, which offers enhanced privacy protection and real-world application.
\subsection{Rehearsal-Free Continual Learning} Early rehearsal-free methods\cite{LWF, zenke2017continual,aljundi2018memory} utilize regularization to alleviate model forgetting. While these approaches demonstrate strong performance in task-incremental continual learning, their effectiveness diminishes in the more challenging class-incremental learning settings. \textit{Generative-based methods}~\cite{choi2021dual,gao2022r,smith2021always} employ deep model inversion to generate images from previous tasks. While these methods perform well in some scenarios, inversion is hampered by slow training speed and high computational costs. Recent studies~\cite{l2p,dualprompt,smith2023coda, zhou2023revisiting, tang2023prompt, gao2024consistent} have incorporated large pre-trained models and prompt learning into the realm of continual learning. Large-scale pre-trained models have demonstrated remarkable generalization performance and robust anti-forgetting capabilities. Therefore, prompt-based methods serve as our primary comparison benchmarks and they will be subsequently elaborated upon in detail.
\subsection{Prompting for Continual Learning} Prompt learning has garnered widespread success in both natural language processing~(NLP)\cite{devlin2018bert,hu2021lora, qin2022deep} and computer vision~(CV)\cite{jia2022visual,tu2023visual, qin2024cross, qin2024noisy}. It leverages prompts to fine-tune large-scale pre-trained models for downstream tasks. Firstly, L2P\cite{l2p} pioneers the integration of prompt learning into the field of continual learning field. It proposes a key-query matching strategy for prompt selection from the prompt pool, followed by the insertion of these prompts into the pre-trained model. The subsequent works are based on this key-query matching strategy. DualPrompt~\cite{dualprompt} employs a set of task-specific prompts and selects prompts based on the closest key-query matching during inference. Furthermore, CODA~\cite{smith2023coda} utilizes key-query matching to assign weights to prompts and can be optimized in an end-to-end fashion. APG~\cite{tang2023prompt} proposes a learnable adaptive prompt generator eases the reliance on intensive pretraining.
S-Prompts\cite{wang2022s} focuses on domain-incremental learning by acquiring distinct prompts for each domain and utilizing KNN to select prompts during inference. In this paper, we thoroughly compared these methods to CIL and DIL protocols and extensively discussed their shortcomings.

\section{Methods}
\subsection{Problem Setting}
The continual learning aims to enable the model to learn non-stationary data from sequential tasks while preserving the knowledge obtained from previous tasks. We define a set of tasks $\mathcal{D}_1, \mathcal{D}_2$,...,$\mathcal{D}_T$, where $\mathcal{D}_t$ = $\{({x_i}^t,{y_i}^t)\}_{i=1}^{N_i}$ is the $t$-th task with ${N_i}$ samples, and ${x_i}^t$ and ${y_i}^t$ is the $i$-th images and its label. The label sets of $\mathcal{D}_t$ are defined as $\mathcal{C}_t$. The objective of rehearsal-free continual learning is to employ a unified model for classifying test images from all previously learned tasks while not storing any old samples or prototypes.

Depending on the specific problems, continual learning can be divided into multiple settings, which differ in challenge and practicality. In this paper, we mainly focus on addressing the challenging \textit{class-incremental learning setting}. In this scenario, the class of each task is non-overlapping, denoted as $\mathcal{C}_1 \cap \mathcal{C}_2 \cdots \cap \mathcal{C}_T = \emptyset$ and the task identity can not be obtained in the inference phase. Furthermore, to assess the robustness and generalization of our approach, we extend our experiments to \textit{domain-incremental continual learning setting}. In this scenario, each task has the same classes but exhibits distinct data distributions, expressed as $\mathcal{C}_1 = \mathcal{C}_2 = \cdots = \mathcal{C}_T$.

\subsection{Framework and Formulation }

\begin{figure*}[t]
\setlength{\abovecaptionskip}{-0.15cm} 
\setlength{\belowcaptionskip}{-0.3cm} 
\begin{center}
    \includegraphics[width=0.9\textwidth]{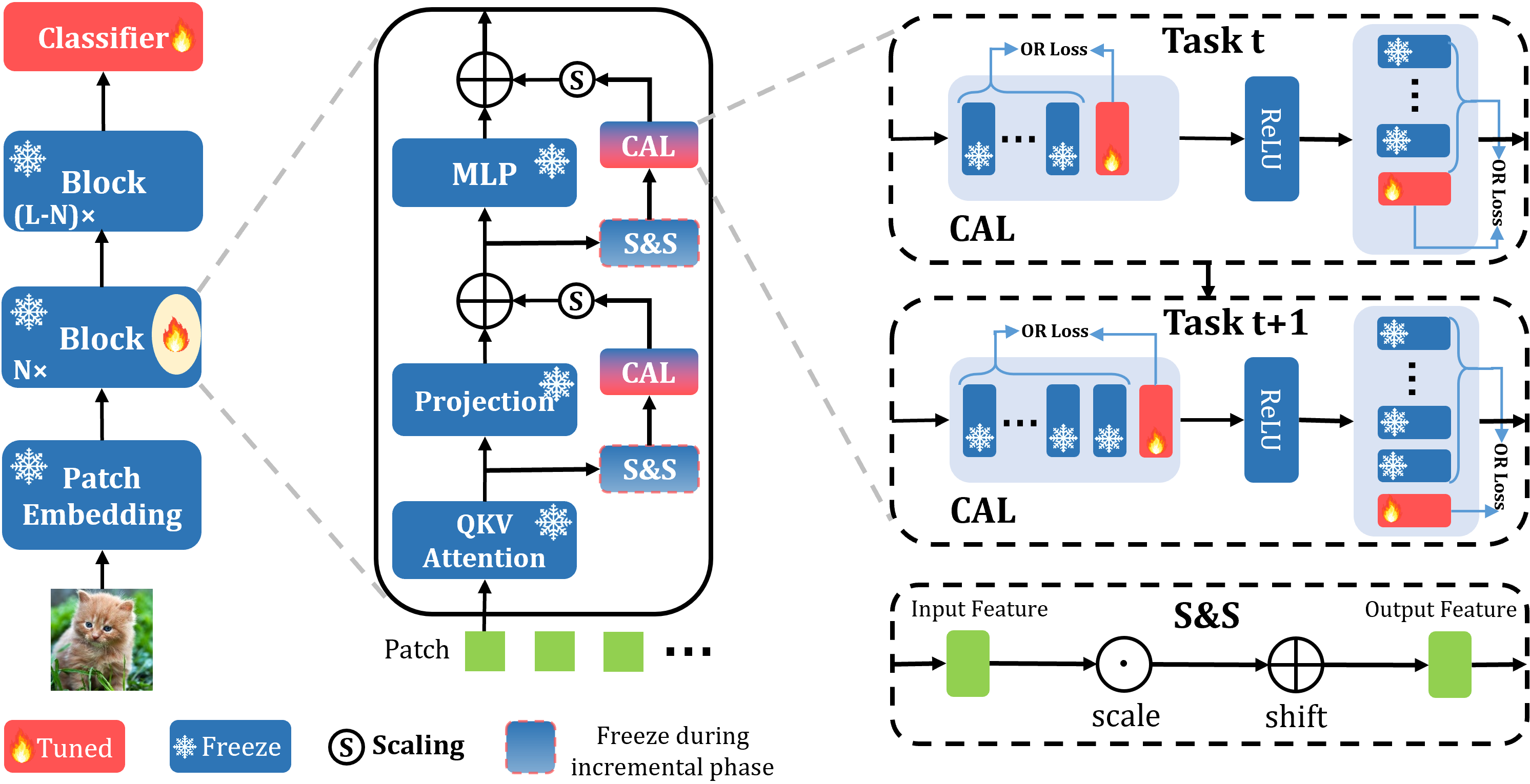}
\end{center}
\caption{The framework of our C-ADA approach. For simplicity, we omit the skip connection and Layernorm in the figure. \emph{or loss} represents the orthogonal loss. We attach our S\&S and CAL, which are in parallel with the projection layer and MLP, to the shallow N layers of ViT and freeze the pre-trained backbone. For each new task, we expand two trainable weights in the CAL to learn the new knowledge and freeze the previous weights. Different from prior works, our approach uses a novel adapter variant to eliminate the necessity for key-query matching. Only trainable weights and classifier parameters are optimized which is \textit{parameter efficient} and no old information~(old images or prototypes) are stored which is \textit{privacy preserving}.}
\label{fig:framework}
\end{figure*}

Figure \ref{fig:framework} illustrates the framework of the Continual Adapter~(C-ADA) approach. The C-ADA incorporates the S\&S module and the plug-and-play Continual Adapter Layer~(CAL) into the pre-trained model, endowing it with robust capabilities for addressing continual learning challenges. Concretely, the S\&S module is exclusively trained before the incremental training to mitigate the domain gap and is subsequently frozen during further training. The CAL comprises a parameter-extensible down-projection, a ReLU activation, and a parameter-extensible up-projection. During incremental learning, the down-projection and up-projection expand the trainable weights to learn the new knowledge and freeze the previous weights to maintain the old knowledge. Moreover, we propose an orthogonal loss to alleviate the conflict between previous weights and trainable weights in each CAL.

The basic backbone consists of a pre-trained vision transformer feature extractor $f = f(\cdot; \theta)$ and a classifier $g = g(\cdot; \phi)$. Given an input image $x \in \mathbb{R}^{H \times W \times C}$ from $\mathcal{D}_t$. The image is reshaped and projected by the patch embedding layer, resulting in the embedding feature $x_e \in \mathbb{R}^{L \times D}$, where $L$ is the token length, and $D$ is the embedding dimension. These embedding features are not only input to the pre-training module within the ViT block but also to the parallel CAL and S\&S modules. The output of the backbone and parallel is added and fed into the next layer. Finally, The output features $x_p$ from the last ViT block are fed into the $g$ to perform classification tasks.

\subsection{Scale and Shift Module}
Due to the significant domain gap between the pre-training dataset and the downstream dataset, our C-ADA approach introduces a flexible and independent module, named scaling and shifting (S\&S), to revise the feature space before incremental training for more efficient fine-tuning~(higher \emph{plasticity}). 

Concretely, the S\&S module includes the trainable parameter $\alpha \in \mathbb{R}^{D}$ and $\beta \in \mathbb{R}^{D}$ as the scale and shift factors, respectively, to learn the domain information of downstream tasks. The S\&S operation is calculated as follows:
\begin{equation}
\begin{aligned}
y = \alpha \odot x + \beta
\end{aligned}
\end{equation}
where $\odot$ is the dot product. We employ the S\&S operation before the incremental training by using the data from the first phase (\emph{i.e.} $\mathcal{D}_1$). Then we freeze the trainable parameters during the incremental phases.

\subsection{Continual Adapter Layer}
\subsubsection{Structure of Continual Adapter Layer}
To address the challenges of efficiency and lightweight, the CAL is designed as a bottleneck module. It consists of a parameter-extensible down-projection, a ReLU layer for non-linear property, and a parameter-extensible up-projection. The down-projection embeds the old and new knowledge into a latent embedding along with the ReLU function, and the up-projection extracts a dimension-fixed knowledge representation from the latent embedding. For each new task, we first initialize and append two learnable linear weights to the down-projection and up-projection, respectively, and freeze the previous weights. Such design ensures that the CAL is extendable for learning new knowledge and stable in maintaining old knowledge, maintaining the input and output feature dimensions unchanged. Assuming the current task t, we assign the weights in the CAL into two types: the frozen weights $\textcolor{blue}{\mathbf{W}^{1}}$,...,$\textcolor{blue}{\mathbf{W}^{(t-1)}}$, and the trainable weight $\textcolor{red}{\mathbf{W}^{t}}$~($\mathbf{W}^{i}$ represents both the $\mathbf{W}_{dp}^i$ and $\mathbf{W}_{up}^i$). The frozen weights effectively preserve the previous knowledge to mitigate catastrophic forgetting, while simultaneously optimizing the trainable parameters to learn new knowledge.

Without loss of generality, we define the input and output of the current layer as $x_i$ and $x_{i}^{'}$, respectively. In task t, we can adapt the input by the CAL as follows:
\begin{equation}
\begin{aligned}
&\mathbf{W}_{dp} = [\textcolor{blue}{\mathbf{W}_{dp}^1},...,\textcolor{blue}{\mathbf{W}_{dp}^{t-1}},\textcolor{red}{\mathbf{W}_{dp}^t}], \\
&\mathbf{W}_{up} = [\textcolor{blue}{\mathbf{W}_{up}^1},...,\textcolor{blue}{\mathbf{W}_{up}^{t-1}},\textcolor{red}{\mathbf{W}_{up}^t}],\\
&\mathbf{W}_{dp}^i \in  \mathbb{R}^{D \times d} \,\,\, \mathbf{W}_{up}^i \in  \mathbb{R}^{d \times D}\\
&x_{i}^{'} =  \text{ReLU}(x_i \cdot \mathbf{W}_{dp}) \cdot \mathbf{W}_{up}
\end{aligned}
\end{equation}
In the next task, we freeze the $\mathbf{W}_{dp}^t$ and $\mathbf{W}_{up}^t$ to maintain the old knowledge and convert two new weights $\mathbf{W}_{dp}^{t+1}$ and $\mathbf{W}_{up}^{t+1}$ to the trainable weights. 

\subsubsection{Parameters of Continual Adapter Layer.} In this paper, we follow the standard RFCL setting where the total task number is known before training. To compare fairly, we adjust the value of $d$ to maintain a consistent number of parameters across different settings. For real-world applications, we can keep the size of $d$ constant and continually expand the specific weights for new tasks.  Remarkably, even when $d$ is set to a minimal value, such as $d$=1, it still exhibits commendable performance. Therefore C-ADA is lightweight, and memory-efficient in real-world applications.

\subsubsection{Position of Continual Adapter Layer.} 
As depicted in \cite{vaswani2017attention,dong2021attention}, the Multi-Head Self-Attention~(MHSA) block employs the attention mechanism to capture global information, while the MLP block prevents the ViT output from degeneration. Both components play pivotal roles in the ViT structure. Consequently, we investigate performing a comprehensive tuning of both the MHSA block and MLP block, instead of exclusively emphasizing one over the other. To fine-tune the MHSA block, we choose to insert the CAL in parallel with the Projection Layer instead of the Attention module. This decision is motivated by the observation that, despite various advanced transformer-based models using different attention mechanisms within the MHSA block~\cite{liu2021Swin,mao2022towards,fan2021multiscale}, they consistently include a projection layer. On the contrary, we insert the CAL in parallel with the entire MLP block to fine-tune the MLP block.

Without loss of generality, we define the input as $x_l$ and initially feed it into the Attention module to obtain $x_a$, which serves as the input to the Projection Layer and CAL$_{1}$:
\begin{equation}
\begin{aligned}
& x_a = \text{Atten}(x_l)=\text{Softmax}\left(\frac{Q \cdot K^{T}}{\sqrt{d}}\right) V, \\
& x_{l1} = \text{Proj}(x_a), \quad x_{l2} = \lambda \cdot \text{CAL}_1(S\&S(x_a)), \\
& x_{l}' = x_{l1} + x_{l2},
\end{aligned}
\end{equation}
Then the output $x_{l}^{'}$ is further sent to the MLP block and CAL$_2$. This process is formally formulated as follows,
\begin{equation}
\begin{aligned}
& x_{l1} = \text{MLP}(x_{l}'), \quad x_{l2} = \lambda \cdot \text{CAL}_2(S\&S(x_{l}')), \\
& x_{l+1} = x_{l1} + x_{l2},
\end{aligned}
\end{equation}
where CAL$_1$ and CAL$_2$ are in parallel with the projection layer and MLP block, respectively. $\text{Proj}$ represents the Projection Layer and $\lambda$ represents a scale factor (typically set to 0.1). Then, the output $x_{l+1}$ is sent to the next layer and repeats the above process.
\subsection{Loss Function}
\subsubsection{Orthogonal Loss.} When training the new task, the learned new knowledge will interfere with old knowledge, potentially resulting in catastrophic forgetting. Thus we propose to solve this problem by initialization and loss function. We apply the Gram-Schmidt process to initialize the trainable parameters at the start of each new task. Orthogonal initialization contributes to the diminishment of the scope when the model parameter is updated, thereby fostering enhanced stability and overall performance. After the first task, we employ the orthogonal loss to keep trainable weights orthogonal to previous weights. 
\begin{equation}
\begin{aligned}
\mathcal{L}_{or} = & \|\textcolor{red}{\mathbf{W}_{dp}^t}^T \cdot \{\textcolor{blue}{\mathbf{W}_{dp}^1},...,\textcolor{blue}{\mathbf{W}_{dp}^{t-1}}\}\|_2  + \|\textcolor{red}{\mathbf{W}_{up}^t} \cdot \{\textcolor{blue}{\mathbf{W}_{up}^1},...,\textcolor{blue}{\mathbf{W}_{up}^{t-1}}\}^T\|_2
\end{aligned}
\end{equation}

where $\|\cdot\|_2$ represent the $L_2$-norm and $T$ represents the matrix transpose. The loss function forces the model to optimize current weight $\mathbf{W}^t$ to be orthogonal to previous weight ${\mathbf{W}^1,...,\mathbf{W}^{t-1}}$. Intuitively, orthogonal vectors exhibit reduced mutual influence. As a result, this design aims to minimize interference between new and old knowledge to reduce catastrophic forgetting. 
\subsubsection{Classification Loss.} Similar to the previous methods, we use the CE loss to optimize the model to learn the new knowledge 
\begin{equation} 
\begin{aligned}
    \mathcal{L}_{ce} = \mathcal{L}\left(g\left(f(\bm{x}; \mathbf{W}_{dp}, \mathbf{W}_{up}, \theta); \phi\right), y\right) 
\end{aligned}
\end{equation}
where $x$ is the input image and $y$ is the corresponding label. $\phi$ is the parameters of the classifier and $\theta$ is the parameters of the pre-train model.

\subsubsection{Total Optimization Objective.} For task t, our full optimization consists of a classification loss $\mathcal{L}_{ce}$ and an orthogonal loss $\mathcal{L}_{or}$
\begin{equation} \label{eq:full_loss}
\begin{aligned}
    \underset{\textcolor{red}{\mathbf{W}_{\text{dp}}^t}, \textcolor{red}{\mathbf{W}_{\text{up}}^t}, \phi}{\operatorname{min}} & \mathcal{L}_{ce} + \delta  \mathcal{L}_{\text{or}}
\end{aligned}
\end{equation}
where $\textcolor{red}{\mathbf{W}_{dp}^t}, \textcolor{red}{\mathbf{W}_{up}^t}$ and $\phi$ are the learnable parameters of task t. We set the $\delta$ to 1 for all the experiments. During each training phase, our models can be optimized in an end-to-end fashion.

\section{Experiments}
In this paper, we mainly aim to solve the \textit{rehearsal-free class-incremental learning} problem. To evaluate our methods, we follow the RFCL settings proposed in previous works\cite{l2p,dualprompt,smith2023coda} and conduct comprehensive experiments. 

Besides, to prove the robustness and generality of our method on other continual learning settings, we conduct experiments compared with the SOTA methods~\cite{wang2022s} on the \textit{domain-incremental learning} settings. Extensive ablation studies prove the effectiveness of our methods.
\subsection{Evaluation Benchmarks}
\subsubsection{Datasets.} We use the Split CIFAR-100\cite{cifar} and Split ImageNet-R\cite{hendrycks2021many} for class-incremental setting, CORe50\cite{lomonaco2017core50} and DomainNet\cite{peng2019moment} for domain-incremental setting.

\textbf{CIFAR-100} is a widely-used benchmark in continual learning, which contains 60000 images of 32 $\times$ 32 size from 100 classes. We split CIFAR-100 into the 10-task benchmark following previous works.

\textbf{ImageNet-R} has 200 classes and includes newly collected data of different styles. It provides a fair and challenging problem setting due to the significant gap between training data and pre-trained data. We split the dataset into the 5-task benchmark, the original 10-task benchmark and the 20-task benchmark.

\textbf{CORe50} has 50 classes from 11 distinct domains. Following the domain-incremental setting, we use 8 domains for~(120000 images) for continual training and the rest of the domains for testing. 

\textbf{DomainNet} is a large dataset with 345 classes and 600000 images. These images are split into 6 domains~(Clipart, Infograph, Paint, Quickdraw, Real, and Sketch) for continual learning like S-Prompts~\cite{wang2022s}.
\subsubsection{Comparing Methods.} We compare our method with several baselines and state-of-the-art~(SOTA) continual learning methods. For a fair comparison, all the methods are based on a ViT-B/16 backbone pre-trained on ImageNet-1K. Under class-incremental settings, we regard L2P\cite{l2p}, DualPrompt\cite{dualprompt}, and CODA\cite{smith2023coda} \footnote{\fontsize{7}{11}\selectfont For the sake of fairness, we primarily selected CODA-P-S with the same parameter count as C-ADA for comparison.} as the main competitors, which are rehearsal-free methods. Under domain-incremental settings, despite S-Prompt\cite{wang2022s} storing the prototypes of previous domains instead of the rehearsal-free method, it is still the main comparison method as the SOTA of domain-incremental learning.


\begin{table*}[t]
\caption{\textbf{Results (\%) on ImageNet-R and CIFAR-100}. The results are all obtained by CODA\cite{smith2023coda} directly. P represents the total number of tasks. UB represents the full fine-tuning result from \cite{dosovitskiy2020vit}. We report $A_N$~(\%), $Param$~(\%) and $FLOPs$~(G) to bring intuitive comparison. C-ADA uses fewer parameters and doubles the training speed, greatly surpassing other methods.}
\centering
\resizebox{0.95\linewidth}{!}{
\begin{tabular}{l|cccccc|cc|c} 
\hline 

  & \multicolumn{6}{c|}{\textbf{ImageNet-R}}   & \multicolumn{2}{c|}{\textbf{CIFAR-100}} & \multicolumn{1}{c}{} \\ \cline{2-9} 

\multicolumn{1}{c|}{\textbf{Methods}} & \multicolumn{2}{c}{\textbf{P=5}} & \multicolumn{2}{c}{\textbf{P=10}} & \multicolumn{2}{c|}{\textbf{P=20}}  & \multicolumn{2}{c|}{\textbf{P=10}}& \multicolumn{1}{c}{\textbf{FLOPs}} \\ \cline{2-9} 

 & $A_N$ ($\uparrow$) & $Param$ ($\downarrow$) & $A_N$ ($\uparrow$) & $Param$ ($\downarrow$) & $A_N$ ($\uparrow$) & $Param$ ($\downarrow$) & $A_N$ ($\uparrow$) & $Param$ ($\downarrow$) &  \\
\hline
UB
& $77.13$ & $100/100$  
& $77.13$ & $100/100$  
& $77.13$ & $100/100$  
& $89.30$ & $100/100$   & - \\ 
\hline
ER (5000)        
& $71.72$ & $100/100 $ 
& $64.43$  & $100/100 $ 
& $52.43$  & $100/100 $ 
& $76.20 $ & $100/100 $ & -
\\  
FT          
& $18.74$ & $100/100 $ 
& $10.12$ & $100/100 $
& $4.75$  & $100/100 $ 
& $9.92 $ & $100/100 $  & -
  \\
FT++          
& $60.42$ & $100/100 $
& $48.93$ & $100/100 $
& $35.98$ & $100/100 $ 
& $49.91$ & $100/100 $  & -\\
LwF.MC       
& $74.56$ & $100/100 $
& $66.73$ & $100/100 $
& $54.05$ & $100/100 $ 
& $64.83$ & $100/100 $  & - \\
L2P    
& $70.83$ & $0.7/100.7 $ 
& $69.29$  & $0.7/100.7$ 
& $65.89$ & $0.7/100.7$  
& $82.50 $ & $0.7/100.7 $  & 35.2 \\
Deep L2P  
& $73.93$  & $9.6/109.6 $ 
& $71.66$ & $9.6/109.6 $
& $68.42$ & $9.6/109.6 $ 
& $84.30 $ & $9.6/109.6$ & 35.2\\
DualPrompt 
& $73.05$  & $0.5/100.5$
& $71.32$  & $0.8/100.8 $ 
& $67.87$  & $1.3/101.3 $ 
 & $83.05 $ & $0.8/100.8$  & 35.2 \\
CODA-P-S
& $75.19$ & $0.7/100.7$
& $73.93$  & $0.7/100.7$
& $70.53$ & $0.7/100.7$ 
& $84.59$ & $0.6/100.6 $ & 35.2\\
\textbf{C-ADA}  
& $\textbf{77.93}$ & $0.7/100.7$
& $\textbf{76.66}$ & $0.7/100.7$
& $\textbf{73.47}$ & $0.7/100.7$
& \textbf{87.18} & $0.6/100.6 $ & 17.6 \\ 
\hline
\end{tabular}
}
\label{tab:imnet-r_main}
\end{table*}
\subsubsection{Implementation Details.} Similar to CODA\cite{smith2023coda}, we use the Adam\cite{kingma2014adam} optimizer with $\beta_1$ = 0.9 and $\beta_2$ = 0.999, and a batch size of 128 images in four GPUs. For every image, we resize it to 224$\times$224 and normalize them to [0,1]. The learning rate is set to $5e^{-5}$ for all the settings with cosine-decaying. For fair comparisons, we set the total middle dimension to 60~(sum of two CALs) and insert the CAL into layers 0-4, which achieves a similar parameter amount to previous works\cite{smith2023coda,dualprompt}. We train the CIFAR-100 for 20 epochs, ImageNet-R for 50 epochs, CORe50 and DomainNet for 10 epoch. During the training phase, we substitute predictions from previous-task logits with negative infinity during the training of a new task. This design leads to a softmax prediction of "0" for these previous task classes and ensures that gradients do not affect the linear heads of past task classes like previous works.\cite{l2p,dualprompt,smith2023coda}

\begin{table}[t]
\caption{\textbf{Results (\%) on CORe50 and DomainNet}. $A_N$ gives the accuracy averaged over tasks and buffer size represents the number of old exemplars stored in the memory buffer. The results are all obtained by S-Prompt\cite{wang2022s} directly. We reproduce the CODA on these settings for further comparison. We ensure that all methods have the same pre-trained model~(ViT-B/16).}
    \begin{minipage}{.48\textwidth}
        \centering 

            \scalebox{0.95}{
\begin{tabular}{c|c|c}
\hline
\rule{0pt}{9pt} Method  & Buffer size & $A_N$ ($\uparrow$) \\
\hline
Upper-Bound & - & 84.01  \\  
\hline
ER   & & $80.10$ \\ 
GDumb        &  & $74.92$   \\  
BiC      &  & $79.28$   \\ 
DER++    & \textbf{50/class} & $79.70$   \\ 
Co$^2$L     &  & $79.75$   \\
DyTox &  & $79.21$ \\  
L2P &  & $81.07$  \\
\hline
EWC      & & $74.82$   \\ 
LwF    & & $75.45$   \\ 
L2P     &\textbf{0/class} & $78.33$   \\
CODA-P-S &  & $85.41$ \\  
S-Prompts &  & $83.13$  \\
\textbf{C-ADA}&  & \textbf{89.45}  \\

\hline
\end{tabular}
\label{table:core50}
            }
    \end{minipage}
    \hfill
    \begin{minipage}{.48\textwidth}
        \centering 
        \scalebox{0.95}{
\begin{tabular}{c|c|c}
\hline
\rule{0pt}{9pt} Method  & Buffer size & $A_N$ ($\uparrow$) \\
\hline
Upper-Bound & - & 63.22  \\  
\hline
DyTox   & \textbf{50/class} & $62.94$ \\ 
\hline
EWC        && $47.62$   \\  
LwF      &  & $49.19$   \\ 
SimCLR    &  & $44.20$   \\ 
BYOL     &  & $49.70$   \\
Barlow Twins & \textbf{0/class} & $48.90$ \\  
Supervised Contrastive &  & $50.90$  \\
L2P      & & $40.15$   \\ 
CODA-P-S &  & $47.56$ \\  
S-Prompts &  & $50.62$  \\
\textbf{C-ADA}&  & \textbf{53.00}  \\

\hline
\end{tabular}
        }
        \label{table:domainNet}
    \end{minipage}
\end{table}

\subsection{Main Results}
\subsubsection{Class-incremental learning.} We report the results on two class-incremental datasets in Table \ref{tab:imnet-r_main} following the previous work\cite{smith2023coda}.

\textbf{ImageNet-R:} We report the results with 5-tasks, 10-tasks, and 20-tasks to evaluate the robustness of the model under different settings in Table \ref{tab:imnet-r_main}. It is worth noting that our method significantly surpasses the previous methods in average accuracy under different settings with as much as \textbf{+2.74\%}, \textbf{+2.73\%} and \textbf{+2.94\%} over CODA, \textbf{+4.88\%}, \textbf{+5.41\%} and \textbf{+5.60\%} over DualPrompt~(second best method). This phenomenon shows that our method is robust to different class increment settings. Moreover, since our method only requires forward propagation one time. We achieved an obvious increase in training speed while improving accuracy. Combining accuracy and training speed brings better practicality in the real world.

\textbf{CIFAR-100:} We note that our approach also brings similar results to the ImageNet-R benchmarks, with an improvement of \textbf{+2.59\%} compared with CODA, \textbf{+4.13\%} compared with DualPrompt.

\subsubsection{Domain-incremental learning.} We report the results on two domain-incremental datasets in Table \ref{table:core50}. Besides, we reproduced CODA on these datasets to bring intuitive comparison.

\textbf{CORe50} and \textbf{DomainNet} are widely used domain dataset for continual learning. Following the setting of S-Prompt\cite{wang2022s}, we divide the baseline methods into two groups, exemplars-based methods\cite{buzzega2020dark,prabhu2020gdumb,BIC,dytox,chaudhry2019tiny,cha2021co2l} and exemplars-free methods\cite{wang2022s}/rehearsal-free methods\cite{l2p,smith2023coda}. Exemplars-based methods store 50 old exemplars per class in the memory buffer. These methods often have better accuracy but are accompanied by the risk of privacy leakage and memory overhead. Our method surpasses the CODA by \textbf{+4.04\%} 
and \textbf{+5.44\%} on CORe50 and DomainNet benchmark. Besides, \textit{it is worth noting that our method even outperforms S-Prompt, which is specifically designed for domain-incremental learning}, with improvements of \textbf{+6.32\%} and \textbf{+2.38\%}. This interesting phenomenon demonstrates the robustness and effectiveness of our method in various continual learning settings.
\subsection{Key Components Ablation}
\begin{figure}[t]
\centering
\begin{minipage}{0.48\textwidth}
\begin{center}
    \includegraphics[width=0.9\textwidth]{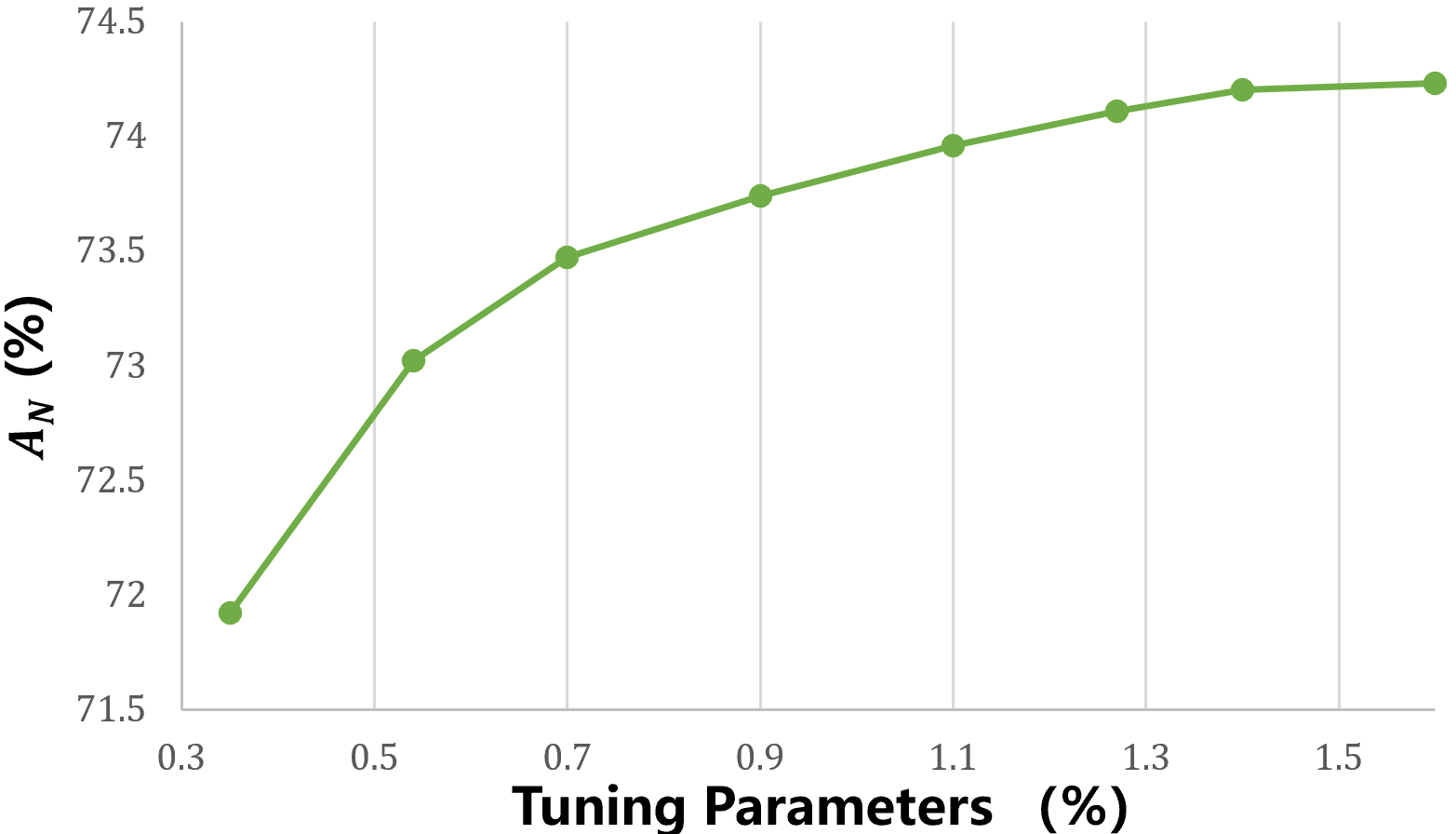}
\end{center}
\caption{\textbf{Average accuracy $A_N$~(\%) vs tuning parameters~(\%)}. We report the results on the 20-tasks ImageNet-R.}
\label{fig:scale}
\end{minipage}
\hfill
\begin{minipage}{0.48\textwidth}
\centering
    \captionsetup{type=table}
        \scalebox{0.9}{
\begin{tabular}{c|c|c} 
\hline
\rule{0pt}{9pt} Method & CIFAR-100 & ImageNet-R \\
\hline
C-ADA             & $87.32 $ & $76.66$   \\ 
\hline
Ablate $S\&S$   & $86.11 $ & $75.93$   \\  
Ablate $\mathcal{L}_{or}$   & $82.89 $ & $72.17$   \\  
Ablate  CAL    & $76.93$ & $64.45$   \\ 

\hline
\end{tabular}}
\caption{\textbf{Ablation Results (\%) on 10-task ImageNet-R and CIFAR-100}. $A_N$ gives the accuracy averaged over tasks. We ablate the key components in turn and report the results.}
\label{tab:ablation}
\end{minipage}
\end{figure}

We conducted ablation experiments to analyze the importance of each component. We incrementally introduced key components on both the 10-tasks ImageNet-R and CIFAR-100 datasets, and the results are presented in Table \ref{tab:ablation}. Firstly, we ablate the S\&S module, the performance has dropped~(\textbf{0.73\%$\downarrow$ on ImageNet-R}). This shows that revising the feature space before training is beneficial to improving subsequent learning ability. Secondly, removing the orthogonal loss leads to severe performance degradation on average accuracy~(\textbf{3.76\%$\downarrow$ on ImageNet-R}). This outcome aligns with our expectations, as ablating the orthogonal loss results in our method lacking a mechanism to prevent forgetting, leading to severe catastrophic forgetting. Thirdly, when we ablated the CAL component, we observed a noticeable drop in performance(\textbf{7.72\%$\downarrow$ on ImageNet-R}). The removal of the CAL implies that our model relies solely on adjusting the classifier to tune the model, which obviously leads to unsatisfactory results, particularly when dealing with the downstream datasets having a significant gap from the pre-trained dataset. 

\subsection{Scaling Trainable Parameters}
We conduct experiments to discuss the relationship between the trainable parameters amount and the performance.

The trainable parameters amount can be adjusted by changing the middle dimension of the CAL. The total number of middle dimensions of the two CALs in each block is chosen from \{20, 40, 60, 80, 100, 120, 140, 160\}, corresponding to \{0.35\%, 0.54\%, 0.71\%, 0.90\%, 1.12\%, 1.27\%, 1.45\%, 1.61\%\} trainable parameters. We report the results on 20-tasks ImageNet-R in Figure \ref{fig:scale}. The figure shows that more trainable parameters contribute to achieving a higher accuracy. Moreover, we can notice that even with the fewer trainable parameters, our model consistently demonstrates superior performance when compared to CODA~(71.92\% $A_N$, 0.35\% $Param$ of C-ADA vs 70.53\% $A_N$, 0.7\% $Param$ of CODA). This fully illustrates that our approach improves the plasticity compared with previous works.
\subsection{Attaching Layer and Position}
Intuitively, each layer of the ViT exhibits distinct feature extraction capabilities\cite{dualprompt}. The impact of tuning different layers may be quite different in CL. Furthermore, each module of the layer plays different roles, which causes attaching CAL to various positions within the layer may yield divergent outcomes. Therefore, it is crucial to explore layers and positions to attach the CAL under the CL settings. Note that the total number of trainable parameters remains consistent across all experiments described below~(middle dim = $d * t$).

\subsubsection{Layer of CAL.} For simplicity, we assume that the attaching layers are contiguous. We report the result in Table \ref{table:layer}.

Firstly, we attach the CAL to all the layers to get a preliminary result~(Layer 0 to Layer 11). Next, we use binary search to analyze the importance of shallow~(Layer 0 to Layer 5) and deep~(Layer 6 to Layer 11) layers. We observe that attaching to shallow layers gets superior results compared to attaching to deeper layers. This phenomenon is consistent with the observation from previous work\cite{dualprompt}. Then we continue to reduce the number of attached layers in shallow layers. Experimental results show that attaching from layer = 0 to layer = 4 performs the best. We use this conclusion to analyze the attached modules.

\subsubsection{Position of CAL.} By analyzing common parts of the ViT architecture, we select three alternative attaching positions: QKV attention layer~($W_q$, $W_k$, $W_v$) in the MHSA block, projection layer in the MHSA block, and the MLP block. We report the result in Table \ref{table:position}.

Firstly, we analyze the effects of only attaching these three positions, respectively. We observe that attaching the CAL to the projection layer has a better effect than in the QKV attention layer. This phenomenon makes us attach the CAL to the projection layer to tune the MHSA block. Then we opt to combine the tuning of different positions. We observe that attaching the CAL to different positions resulted in certain improvements~(about 0.4\%). Interestingly, it seems to have little to do with how to assign the middle dimensions. This shows to some extent that our method is robust.

\begin{table}[t]
    \begin{minipage}{.48\textwidth}
     \caption{\textbf{Results (\%) on ImageNet-R 10 tasks}. We analyze the difference of attaching to different layers. We keep the total tunable parameters constant by modifying the middle dimension,}
        \centering 
            \scalebox{0.9}{
\begin{tabular}{c|c|c}
\hline
\rule{0pt}{9pt} Attach Layer & middle dim & $A_N$ ($\uparrow$) \\
\hline
Layer 0 $\sim$ Layer 11 & 30 & $75.23$ \\ 
Layer 6 $\sim$ Layer 11  & 50      & $72.51$   \\ 
Layer 0 $\sim$ Layer 5   & 50   & $76.21$   \\  
Layer 1 $\sim$ Layer 5   & 60    & $76.29$   \\  
Layer 0 $\sim$ Layer 4   & 60   & \textbf{76.66}   \\  
Layer 2 $\sim$ Layer 5   & 80   & $75.72$   \\   
Layer 0 $\sim$ Layer 3  & 80   & $75.75$   \\  

\hline
\end{tabular}
            }
                        \label{table:layer}
    \end{minipage}
    \hfill
    \begin{minipage}{.48\textwidth}
            \caption{\textbf{Results (\%) on ImageNet-R 10 tasks}. We fix the attaching layers~(layer 0 to layer 4) to analyze two aspects: the attaching positions and the middle dimension of each CAL.}
        \centering 
        \scalebox{0.9}{
\begin{tabular}{c|c|c}
\hline
\rule{0pt}{9pt} Attach Position & middle dim & $A_N$ ($\uparrow$) \\
\hline
QKV Attention & 60  & 76.04 \\ 
MLP & 60  & 76.23 \\ 
Proj & 60  & 76.25 \\ 
Proj + MLP & 20 + 40 & 76.52 \\ 
Proj + MLP& 30 + 30  & 76.57 \\ 
Proj + MLP & 50 + 10 & 76.51 \\ 
Proj + MLP & 10 + 50  & 76.61 \\ 
Proj + MLP & 40 + 20  & \textbf{76.66} \\ 
 
\hline
\end{tabular}
        }
                \label{table:position}
    \end{minipage}
\end{table}

\subsection{Extra Evaluation Metrics}
Although we emphasize that $A_N$, $Param$, and $FLOPs$ are the more important metrics, there are still some common metrics in the CL field.

\textbf{$F_N$}: Average forgetting metrics, represent the drop in task performance averaged over N tasks.

\textbf{$bwt$}: Backward transfer, assesses the quality of representation.

\textbf{$fwt$}: Forward transfer, represents the influence of existing knowledge on the performance of subsequent concepts.

In Table \ref{tab:forget}, we can see that across various metrics, C-ADA demonstrates superior performance. Notably, in terms of $fwt$, which represents the capacity to learn new knowledge, it significantly outperforms the other methods. This emphasizes the effectiveness of C-ADA in learning new knowledge.

\begin{table*}[h]
\caption{\textbf{Results (\%) on ImageNet-R}. The $F_N$ results are all obtained by CODA\cite{smith2023coda} directly. We reproduce the $bwt$ and $fwt$ metrics for further comparison.}
\centering
\scalebox{0.95}{
\begin{tabular}{l|ccc|c|c} 
\hline 

\multicolumn{1}{c|}{\textbf{Methods}} & \multicolumn{1}{c}{\textbf{P=5}} & \multicolumn{1}{c}{\textbf{P=10}} & \multicolumn{1}{c|}{\textbf{P=20}}  & \multicolumn{1}{c|}{\textbf{P=10}}& \multicolumn{1}{c}{\textbf{P=10}}\\ \cline{2-6} 

& $F_N$ ($\downarrow$) & $F_N$ ($\downarrow$) & $F_N$ ($\downarrow$) & $bwt  $($\uparrow$) & $fwt$($\uparrow$)\\
\hline
L2P      
& $3.36\pm0.18$ & $2.03\pm0.19$ 
& $1.24\pm0.14$  & $-3.69\pm0.07 $ 
& $-0.32 \pm 0.04$ 
\\  
DualPrompt          
& $2.64 \pm0.17$ & $1.71 \pm 0.24$ 
& $1.07 \pm 0.14$ & $-3.48  \pm 0.06$
& $-0.09 \pm 0.04$  
  \\
CODA         
& $2.65 \pm 0.15$ & $1.60 \pm 0.20 $
& $\textbf{1.00} \pm 0.15$ & $-3.24  \pm 0.06 $
& $0.01 \pm 0.03$ \\
C-ADA       
& $\textbf{2.55} \pm 0.14$ & $\textbf{1.58} \pm 0.18 $
& $1.02 \pm 0.12$ & $\textbf{-3.12} \pm 0.05 $
& $\textbf{0.34} \pm 0.03$ \\

\hline
\end{tabular}
}
\label{tab:forget}
\end{table*}

\section{Conclusion}
In this paper, we present a novel adapter variant, named Continual Adapter~(C-ADA), for RFCL. Our approach assembles a parameter-extensible adapter layer~(CAL) and an S\&S module in parallel with the pre-trained model. Importantly, C-ADA eliminates the necessity for key-query matching to achieve better plasticity and efficient training. We set a new SOTA on well-established benchmarks of RFCL and outperform the mainstream methods of DIL. This underscores the generality and robustness of our approach across various continual learning settings.

\section*{Acknowledgments}
This work was funded by the National Natural Science Foundation of China under Grant No.U21B2048 and No.62302382 and Shenzhen Key Technical Projects under Grant CJGJZD2022051714160501 and the CAAI-MindSpore Open Fund, developed on OpenI Community.

%
%
\bibliographystyle{splncs04}
\bibliography{main}
\end{document}